\documentclass[11pt]{article}

\usepackage[]{acl}

\usepackage{times}
\usepackage{latexsym}
\usepackage{CJKutf8}

\usepackage[T1]{fontenc}

\usepackage[utf8]{inputenc}

\usepackage{microtype}

\usepackage{inconsolata}

\usepackage{booktabs} 
\usepackage{bm}
\usepackage{multirow}
\usepackage{xspace}
\usepackage[tight,footnotesize]{subfigure}
\usepackage{algorithm}  
\usepackage{amsmath}
\usepackage{enumitem}
\usepackage{algorithmic}
\usepackage{balance}
\usepackage{cancel}
\usepackage{graphicx}
\usepackage{enumerate}
\usepackage{amssymb}
\usepackage{bbding}
\usepackage{float}
\usepackage{verbatimbox}

\usepackage{colortbl}

\newcommand{\paratitle}[1]{\vspace{1.5ex}\noindent\textbf{#1}}
\newcommand{\ie}{\emph{i.e.,}\xspace}

\newcommand{\eg}{\emph{e.g.,}\xspace}

\newcommand{\wrt}{\emph{w.r.t.}\xspace}
\newcommand{\st}{\emph{s.t.}\xspace}
\newcommand{\benchmark}{\textsc{PolyEval}}
\newcommand{\model}{\textsc{PolyRag}}

\title{\textsc{PolyRag}: Integrating Polyviews into Retrieval-Augmented Generation for Medical Applications}

\author{Chunjing Gan\quad Dan Yang\quad Binbin Hu\quad Ziqi Liu \quad Yue Shen \\ \bf Zhiqiang Zhang\quad Jian Wang\quad Jun Zhou\footnotemark[2] \\
        Ant Group \\
        jun.zhoujun@antgroup.com
        }

\begin{document}
\maketitle

\begin{abstract}
Large language models (LLMs) have become a disruptive force in the industry, 
introducing unprecedented capabilities in natural language processing, logical reasoning and so on. 
However, the challenges of knowledge updates and hallucination issues have limited the application of LLMs in medical scenarios, where retrieval-augmented generation (RAG) can offer significant assistance.
Nevertheless, existing retrieve-then-read approaches generally digest the retrieved documents, without considering the timeliness, authoritativeness and commonality of retrieval.
We argue that these approaches can be suboptimal, especially in real-world applications where information from different sources might conflict with each other and even information from the same source in different time scale might be different, and totally relying on this would deteriorate the performance of RAG approaches.
We propose {\model} that carefully incorporate judges from different perspectives and finally integrate the polyviews for retrieval augmented generation in medical applications.
Due to the scarcity of real-world benchmarks for evaluation, to bridge the gap we propose {\benchmark}, a benchmark consists of queries and documents collected from real-world medical scenarios (including medical policy, hospital \& doctor  inquiry and healthcare) with multiple tagging (\eg timeliness, authoritativeness) on them.
Extensive experiments and analysis on {\benchmark} have demonstrated the superiority of {\model}\footnote{We will release the data of {\benchmark} soon.}.
\end{abstract}

\section{Introduction}
Recently, large language models (LLMs) such as GPT4 \cite{gpt4_2023}, Llama3 \cite{llama32024}, Qwen \cite{qwen25_2024}, Deepseek-R1 \cite{deepseekr12024} have become a disruptive force in the industry, which introduces marvelous capabilities in natural language processing \cite{whennottotrustlms2023}, logical reasoning \cite{multilogieval2024}, multi-modal processing \cite{multimodalllm2024} and so on. 
However, the heavy costs of knowledge updates \cite{replug2024} and the longstanding hallucination issues \cite{ragsurvey2023} have limited the application of LLMs in medical scenarios where incorrect answers may result in severe consequences, in this case retrieval-augmented generation (RAG) can be of help.
Nevertheless, existing retrieve-then-read approaches generally directly digest the documents from the retrieval stages \cite{selfrag2024}, without considering other perspectives such as timeliness, authoritativeness and commonality of retrieval.

\begin{figure}[!t]
\centering
\includegraphics[width=1.0\columnwidth]{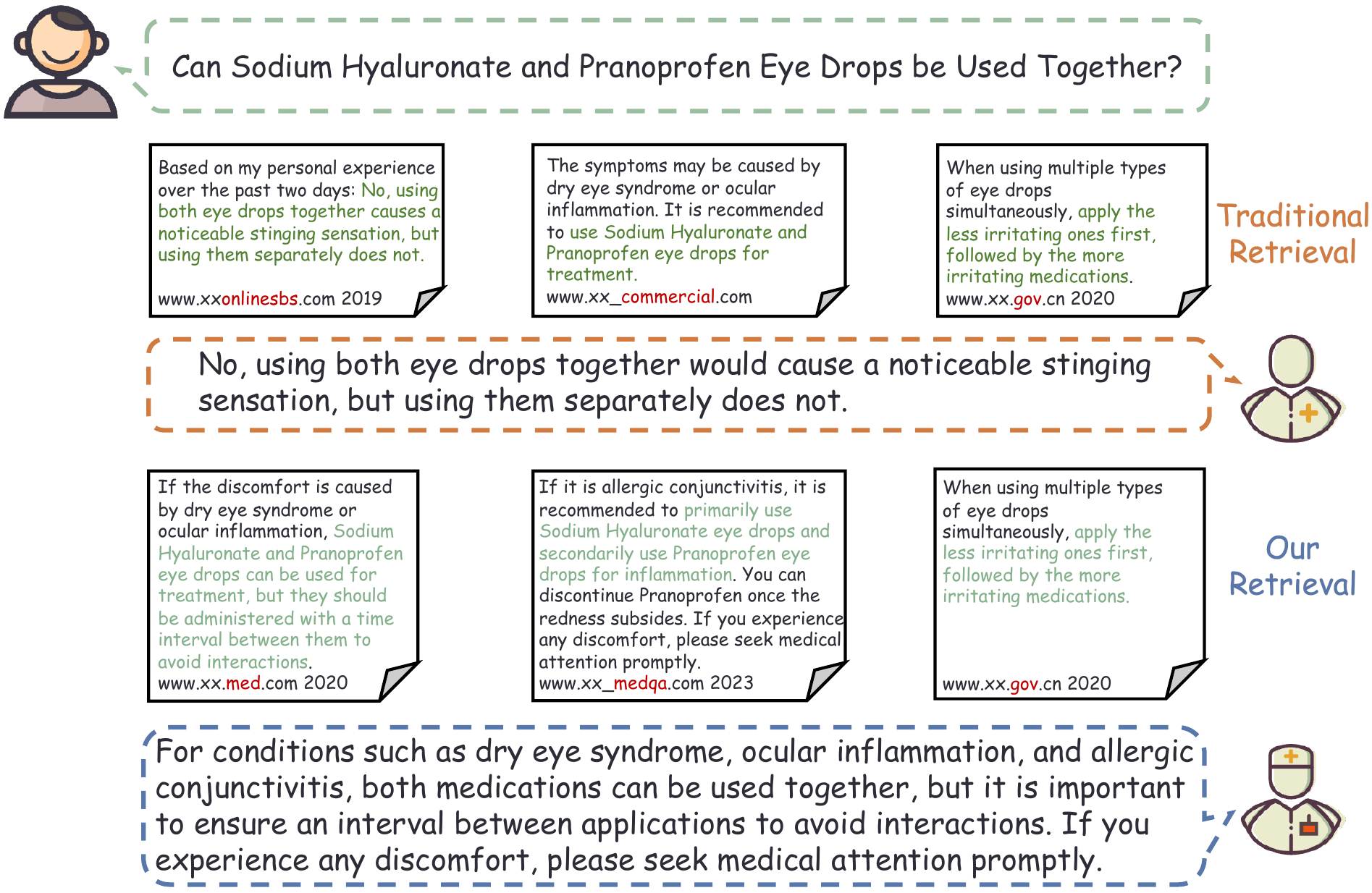}
\caption[Caption for examples]{A toy example illustrating the difference between traditional retrieval and our retrieval strategy, where beyond relevance of a document, we also takes other perspectives such as its authoritativeness into consideration.}
\label{fig:intro_example}
\end{figure}

Here, we argue that oftentimes these approaches can be suboptimal, especially in real-world applications (\eg medical applications) where not only information from different sources with respect to the same fact might conflict with each other but also information from the same source in different time scale might be different, and directly relying on them for generation would deteriorate the performance of RAG approaches.
As the toy example shown in Figure \ref{fig:intro_example}, when a user types in the query ``Can Sodium Hyaluronate and Pranoprofen Eye Drops be Used Together?'', a traditional RAG system would search and rank documents according to its relevance to the query \cite{replug2024}. 
Though the retrieved documents comes from non-authoritative websites and even contradicts with each other such that the LLM used for generation struggles in incorporating the retrieved information, \eg the first document just states they cannot be used together but separately without further context, the second document states they can be used for treating dry eye syndrome or ocular inflammation while the third document states the order of usage, however, various discussions held on this topic do not result in a definitive conclusion which finally hinders its effectiveness for question answering. Not to mention that for some complex queries that contains multiple factors, the top retrieved documents may only contains facts focusing on one factor and ignores documents with respect to other factors, which would severely hinder the performance.

Given the above limitations in current approaches, instead of solely relying on the relevance of documents for generation, we aim to integrate polyviews (\ie multiple views \wrt retrieval such as utility, complement, authoritativeness, timeliness and composibility) into consideration so as to promote its application in medical applications. However, the solution is quite non-trivial, which needs to tackle the following challenges: 
(\textbf{C1}) With multiple views to evaluate, how to measure them and its feasibility in real-world applications remains unknown.
(\textbf{C2}) With the evaluated results of multiple views, in real-world applications what we needed is actually an integrated scoring strategy that comprehensively evaluates each view, how to develop a reasonable and applicable ranking strategy to combine the precedent views remains unanswered.
(\textbf{C3}) The lack of benchmark data that evaluates the retrieval performance of a model from multiple views prohibits us from further developing our model.

To this end, we propose {\model}.
In particular, given that there are many available small but performant models, we carefully allocate storage to make this modeling feasible. (\textbf{C1})
To comprehensively integrate the results of each view, we transform the modeling of ranking strategy to a multi-reward problem and find the mixture of different views. (\textbf{C2})
Due to the scarcity of real-world benchmarks for evaluation, to bridge the gap we propose {\benchmark}, which is a benchmark consists of queries and documents collected from real-world healthcare scenarios (including medical policy, hospital recommendation and medical care) with multiple tagging (\eg timeliness, authoritativeness) on them (\textbf{C3}).
With the polyviews gained from the precedent procedures, we apply the retrieved top-k documents and call an LLM for knowledge-augmented generation. We evaluate the proposed {\model} on multiple tasks and extensive experiments and analysis on {\benchmark} have demonstrated the superiority of the proposed {\model}.

\section{Related Work}\label{sec:rel}
Retrieval-augmented generation (RAG) approaches which empower large language models (LLMs) with additional knowledge and henceforth less need for additional training \cite{rag_survey3_2023,rag_survey1_2024,rag_survey2_2024,sfrrag2024} have been successfully applied to various fields\cite{chatgptsearch2023,raft2024,generatethenground2024,cpr2024,longrag2024} including recommender systems\cite{RAGSys_recommend_2024,ramo_recommend_2024,federated_rec-2024}, question answering\cite{selfrag2024,richrag2025} and so on. 
Among them, question answering in medical applications poses significant challenges due to their high professionalism and low fault-tolerance characteristics. 
Existing approaches for medical-based RAG have been studying additional knowledge acquisition\cite{medcpt2023,med_textbook_24}, query construction\cite{ominirag2024,rationalrag2024}, complex retrieval strategy\cite{medicalgraphrag2024,iterativequery2024,mbarag2025}, complex reasoning\cite{planrag2025,med_cok_2024,medex2025} and so on with focus on better retrieval strategy from external source and better utilization strategy when employ LLMs for answer generation.

\paratitle{Open issues.}
Few research works consider multiple perspectives of the retrieval results and in this work we delve into a direction that can be directly integrated into these existing pipelines where we investigate on how to incorporate retrieval from polyviews for downstream tasks and henceforth promoting retrieval.

\begin{figure*}[t]
    \centering
    \includegraphics[width=1\textwidth]{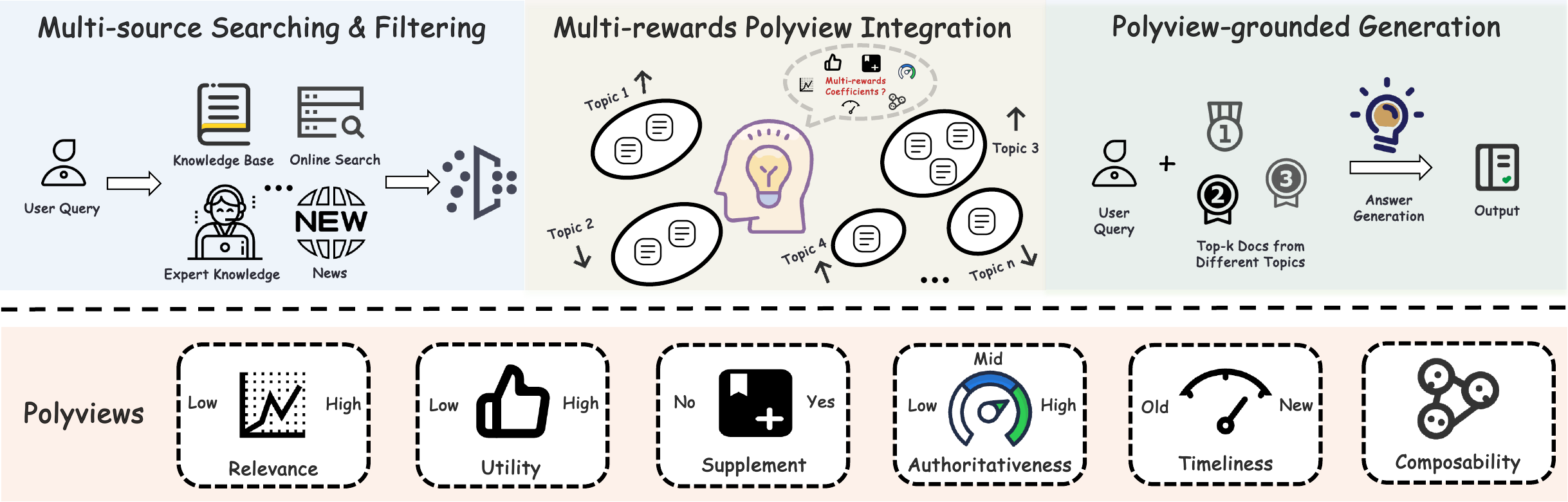}
    \caption{The proposed {\model} framework.}
    \label{fig:framework}
\end{figure*}

\section{The Proposed Approach}\label{sec:model}
\subsection{Overview}
The task of retrieving top critical documents from previous searching and filtering stage is equivalent to comprehensively evaluate the input documents, \ie evaluate the retrieved document from $m$ polyviews $\mathcal{V}$. 
For simplicity, with the assumption that multiple polyviews are independent, given an input query $q$, a document $d$ ($d \in \mathcal{D} = \{d_1, d_2, ..., d_n\}$), where we first evaluate each document independently as follows:
\begin{equation}
\mathbf{P}(d_j \mid \mathcal{V}_{1,j}, \ldots, \mathcal{V}_{m,j}) = \prod_{i=1}^{m} \left( \mathbf{P}(d_j \mid \mathcal{V}_{i,j}) \right)^{w_i}, 
\end{equation}
where $\mathcal{V}_{i,j}$, $w_i$ denote the $j$th document evaluate from the $i$th view regarding the input query $q$, the weight of $i$th view respectively. Given some pre-defined constraints $\mathbb{C}$, we can obtain top-ranking documents $\mathcal{D}_{\text{Top}}$:
\begin{equation}
\mathcal{D}_{\text{Top}} =  \left\{d \in \mathcal{D} \quad \st \quad \mathbb{C} \right\} \\
\end{equation}
In this work, we propose {\model}, as shown in Figure \ref{fig:framework}. With the multi-source searching and filtering results, {\model} firstly embrace varied views for evaluation of each retrieved document (detailed in Section \ref{sec:model_polyviews}) and further pursuing integrated polyviews via a multi-rewards based view-mixture mechanism (detailed in Section \ref{sec:model_merge}), then incorporating the derived polyview-grounded knowledge for answer generation (detailed in Section \ref{sec:model_generation}). 

\subsection{Through Different Lenses: A Document Evaluated via Polyviews}\label{sec:model_polyviews}
In this paper, we pre-define $6$ polyviews, \ie \textit{Relevance} ($\mathcal{R}$), \textit{Utility} ($\mathcal{U}$), \textit{Supplement} ($\mathcal{S}$), \textit{Authoritativeness} ($\mathcal{A}$), \textit{Timeliness} ($\mathcal{T}$), \textit{Composibility} ($\mathcal{C}$, which is used as a retrieval constraint) and detail the estimation of each in the following.

\textit{Relevance} View is a case of symmetric retrieval, which is designed to be direction-agnostic. With an off-the-shelf model $\mathbf{E}$ (which could be a dense retriever followed by a predefined metric $\mathcal{M}$ such as cosine similarity for simplicity or large language models by designing instruction $\mathbb{INS_{\mathcal{R}}}$), we can efficiently obtain the \textit{Relevance} score between the query and document as follows:
\begin{flalign} 
    \mathcal{R}(q,d) &=\left\{\begin{array}{ll}
    \mathbf{P}_\textit{LLM}(d|q,\mathbb{INS_{\mathcal{R}}}), & \text { with LLM;}\\
    \mathcal{M}(\mathbf{E}(q),\mathbf{E}(d)), & \text{otherwise.} 
    \end{array}\right.
\end{flalign}\noindent
However, \textit{Relevance} cannot guarantee usefulness, where we introduce asymmetric retrieval \ie \textit{Utility} View that measures the extent that one document is useful for assisting an LLM to answer the given query, which is modelled by the probability of generating correct answer $a$ with a specific LLM, by designing an appropriate instruction $\mathbb{INS_{\mathcal{U}}}$ to guide the LLM, we can calculate the \textit{Utility} of a document \wrt the input query as follows:
\begin{equation}
    \mathcal{U}(d|q,a) = \mathbf{P}_\textit{LLM}(a|q,d,\mathbb{INS_{\mathcal{U}}}).
\end{equation}
Oftentimes there are documents that do not directly answer the query but they can provide additional knowledge, background information, or alternatives that help users to make more informed decisions or better understand the treatment process, where we define it as the \textit{Supplement} View of a document \wrt the input query, with a carefully designed $\mathbb{INS_{\mathcal{S}}}$ to guide the LLM for estimating \textit{Supplement}, we can formalize it as follows:
\begin{equation}
    \mathbf{S}(d|q) = \mathbf{P}_\textit{LLM}(d|q,\mathbb{INS_{\mathcal{S}}}).
\end{equation}
Besides, given the retrieved documents from previous stage, it is of great significance to take into account the \textit{Authoritativeness} and \textit{Timeliness} Views of them, since that for scenarios with strong professionalism, \ie medical applications in our case, medical treatments recommended by different sources, such as professional doctors and individual accounts, can vary greatly. Additionally, medical policies and practices may evolve over time. Therefore, keeping track of these two dimensions is crucial and here we denote these two dimensions of document $d$ as $\mathcal{A}(d)$ and $\mathcal{T}(d)$\footnote{We approximate $\mathcal{A}(d)$ via $\mathcal{A}(d_{source})$ for simplicity to reduce tagging costs, where the $\mathcal{A}(d_{source})$ is annotated by human annotators. For $\mathcal{T}(d)$, we employ efficient tool for date extraction.}. 
Moreover, the retrieved documents might cover multiple topics \wrt the input query and directly ranking may lead to top documents focusing on partial topics, therefore, we introduce \textit{Composibility} View to account for the difference of topics among them, where the topic of each document can be assigned via an LLM or clustering algorithms to maximize its assigning probability as follows:
\begin{equation}
        \mathcal{C}_d = \arg\max_{k} \mathbf{P}(C_k|d_i) \approx \arg\max_{k} \mathbf{P}(d_i|C_k)\mathbf{P}(C_k).
\end{equation}
  
\subsection{A Cord of Three Strands is Not Quickly Broken: Multi-rewards Boosted Polyview Integration}\label{sec:model_merge}
Given the polyview evaluation results, to efficiently incorporate them for downstream generation, motivated by the idea and marvelous performance in simple rewards-driven reinforcement learning, here we model the integration as multi-rewards integration to obtain an effective mixture of polyviews, for each document $d$ from $\mathcal{D}$, the polyview integration score can be formalized as follows: 
\begin{equation}
y_d = \alpha_1 d_\mathcal{R} + \alpha_2 d_\mathcal{U} + \alpha_3 d_\mathcal{S}  + \alpha_4 d_\mathcal{A} + \alpha_5 d_\mathcal{T},
\end{equation}
where the coefficients can be obtained either by expertise designation or learning from models. 
With the polyview integrated score, we can obtain the top-ranking documents $\mathcal{D}_{\text{Top}}$ under the \textit{Composibility} constraints so that top-ranking documents can cover different topics \wrt the input query: 
\begin{equation}
 \left\|\mathcal{C}_d, d \in \mathcal{D}_{\text{Top}} \right\| \approx \left\|\mathcal{C}_d, d \in \mathcal{D} \right\|.
\end{equation}

\subsection{Polyview-grounded Generation}\label{sec:model_generation}
With the input query $q$ and the polyview-grounded knowledge $\mathcal{P}$ that scatter across different topics related to the query, we can directly call an LLM (it 
 can also be fine-tuned in a supervised manner), where its knowledge-augmented generation output $o$ can be formalized as follows:
\begin{equation}
o^* = \underset{o}{\arg\max} \; \mathbf{P}(o|q,\mathcal{P}),
\end{equation}
where $\mathbf{P}(o|q,\mathcal{P})$ is the probability of the output $o$ given the query $q$ and the external documents $\mathcal{P}$, and $\arg\max$ denotes the argument of the maximum, i.e., the answer $o$ for which $\mathbf{P}(o|q, \mathcal{P})$ is maximized.


\begin{figure*}
    \centering
    \includegraphics[width=1\textwidth]{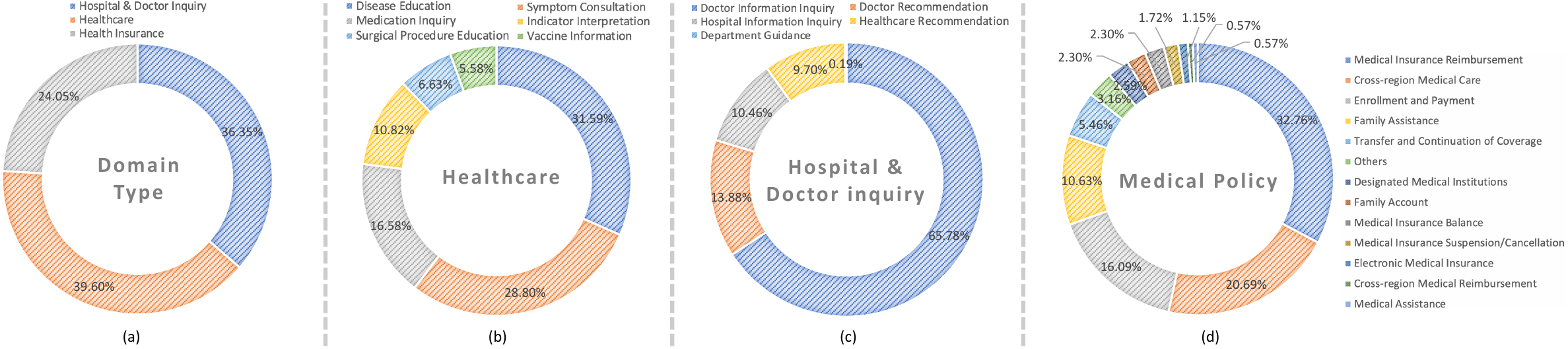}
    \caption{Data distribution of {\benchmark}, where Figure (a) denotes the domain type distribution and Figure (b-d) denote the query intent distribution within each domain.}
    \label{fig:benchmark_distribution}
\end{figure*}

\section{Benchmark}\label{sec:benchmark}
We will first describe the characteristics of {\benchmark} and then delve into its creation process.

\subsection{Characteristics}
To ensure that {\benchmark} can be representatives of real-world medical application user cases, we carefully design it to be diverse in the following three perspectives.
\begin{itemize}[leftmargin=*]
    \item \textbf{Domain Type}: {\benchmark} contains questions from diverse domains including Medical Policy, Healthcare, Hospital \& Doctor Inquiry in order to cover different real-world medical scenarios.
    \item \textbf{Query Intent}: Given questions in each domain, they encompass various types of real user intents, \eg Medical Insurance Balance in Medical Policy domain, Medication Inquiry in Healthcare domain in order to comprehensively represent user needs.
    \item \textbf{Annotation Dimension}: Given a query, for each retrieved document, it is annotated with tags on \textit{relevance}, \textit{complement}, \textit{utility}, \textit{publish date} and \textit{authority level}.
\end{itemize}

\subsection{Benchmark Creation}
\subsubsection{Data Collection}
We collect $1,447$ real-world user queries from a large-scale online platform that offers medical-related services in China, where its distribution of domain type and query intent is illustrated in Figure \ref{fig:benchmark_distribution}\footnote{Due to space limit, we only used the first-level categories when drawing the query intent distribution. In total, there are 40 labels when considering the second-level categories.}. Given each query, we perform multi-source (including expert knowledge, online search engine, knowledge bases and news) documents searching to find relevant documents for annotation. In sum, we have collected $21,276$ documents, making $14.7$ documents for each query on average.

\subsubsection{Annotation Details}
Overall, {\benchmark} is annotated by human annotators or automated tools. For each query and its associated documents, three highly-skilled annotators who have received professional medical training are involved for document \textit{relevance}, \textit{complement} and \textit{utility} annotation and the annotation result is ``accepted'' if at least two annotators reach an agreement unless it is ``rejected''. For \textit{authority level} of document, we approximate it via the \textit{authority level} of its source, \ie we firstly collect abundant information from multiple sources such as medical-related websites and random sample information from them, and then ask human annotators to judge the overall authority of these sources and finally come up with the \textit{authority level}. For \textit{publish date} of document, we employ efficient automated tools for date extraction.

\section{Experiments}\label{sec:exp}

\subsection{Experimental Setup}\label{sec:exp_setup}
\subsubsection{Tasks}
We evaluate our proposed {\model} and multiple baselines for retrieval and generation on {\benchmark} and evaluate the performance of retrieval via  metrics HIT, NDCG, and generation via judge model (\eg GPT 4). 
To better demonstrate the difference between domains in {\model}, we denote data of domain Healthcare, Hospital \& Doctor Inquiry, Medical Policy as $\mathbb{CARE}$, $\mathbb{INQUIRY}$ and $\mathbb{POLICY}$ respectively for simplicity.

\subsubsection{Baselines}

We evaluate models augmented with retrieval via publicly available retrieval model including BM25, GTE \cite{gte_2023}, BGE-M3 \cite{bgem3_2024}, jina embedding v3 \cite{jinaembeddingsv3_2024}.
With the top-$k$ retrieved documents, we directly call strong publicly available pre-trained LLMs,
Qwen2.5$_{\textsc{7B},\textsc{14B},\textsc{32B}}$~\cite{qwen25_2024} for generation.

\subsubsection{Training, Generation and Evaluation Details.}\label{sec:exp_setup_settings}
Our training data includes randomly sampled <query,document,label> triples (which are excluded from {\benchmark})\footnote{For \textit{Relevance} and \textit{Supplement} evaluation, the label is binary, 
\ie 0 or 1 while for \textit{Utility} evaluation the label is a float number generated by a powerful LLM.} from a large-scale medical service platform in China to train our model for evaluating polyviews. All experiments are conducted using 4 NVIDIA A100 GPUs.
For \textit{Relevance} and \textit{Supplement} evaluation, we utilize open-source Llama Factory\footnote{https://github.com/hiyouga/LLaMA-Factory} to finetune small-scale Qwen2.5$_{\textsc{1.5B}}$ and adopt Lora tuning for 5 epoch with a learning rate of 5e-5, a batch size of 4 and a cosine learning rate scheduler.
As for \textit{Utility} evaluation, we incorporate BGE-M3 owing to its superior performance in a variety of benchmark leaderboards and distill the marvelous power of LLM in evaluating utility into it, where $\mathcal{M}(\cdot)$ is defined as cosine similarity. We train the utility model for 5 epochs with a learning rate of 1e-5, a batch size of 16 for each device, a warm-up ratio of 0.2, the passage window size of 50 and the temperature parameter $\tau$ set to 0.05 following \cite{metrag_2024}.
For \textit{Composibility} evaluation, we borrow the embedding from \textit{Utility} and conduct clustering via DBSCAN \cite{dbscan_1996}.
For all generation tasks, we utilize vLLM \cite{vllm} for inference speed-up and set the temperature to 0 for reproducibility and max token parameter to 1. 
We set $[\alpha_1,\alpha_2,\alpha_3,\alpha_4,\alpha_5]$ is set to [0.35, 0.35, 0.1, 0.1, 0.1] for $\mathbb{INQUIRY}$ and $\mathbb{POLICY}$ and [0.35, 0.35, 0.1, 0.2, 0.0] for $\mathbb{CARE}$ for simplicity.
For generation evaluation, we directly call private commercial LLM GPT4 to conduct answer statement generation and the judgement (\ie circumstances \underline{c}orrect, \underline{i}ncorrect and \underline{n}ot mentioned) between answer statement and ground truth and $\mathbb{N}_c$, $\mathbb{R}_c$ denote the count and ratio of the given circumstance $c$. Finally, we have listed all prompt templates in the Appendix.

\subsection{Results and Analysis}

\begin{table}[t]\centering
\scriptsize
\setlength{\tabcolsep}{1.75mm}{

\caption{Overall retrieval performance (\%) evaluation on {\benchmark}, here $k$ is set to 3 for simplicity. 
} 
\label{tab:overall_retrieval_result_top3}
\begin{tabular}{c|c|c|c|c|c|c}
\toprule
\multirow{2}{*}{Retrieval}
& \multicolumn{2}{c|}{$\mathbb{CARE}$} 
& \multicolumn{2}{c|}{$\mathbb{INQUIRY}$} 
& \multicolumn{2}{c}{$\mathbb{POLICY}$} 
\\    
\cmidrule{2-7}
 {}
& {HIT} & {NDCG} 
& {HIT} & {NDCG}
& {HIT} & {NDCG}\\ 
\midrule 
{BM25}
& {26.6} & {26.6} 
& {22.3} & {22.7} 
& {28.2} & {28.6} \\ 
\midrule
{GTE}
& {38.7} & {39.4} 
& {31.1} & {31.7} 
& {31.4} & {32.0} \\ 
\midrule
{BGE-M3}
& {40.0} & {40.8} 
& {34.8} & {35.7} 
& {33.7} & {34.6} \\ 
\midrule
{jina}
& {42.8} & {43.7} 
& {33.5} & {34.7} 
& {35.9} & {36.9} \\ 
\midrule
{{\model} } 
& {\textbf{47.1}} & {\textbf{48.3}} 
& {\textbf{38.1}} & {\textbf{39.1}} 
& {\textbf{42.8}} & {\textbf{44.5}} \\ 
\bottomrule
\end{tabular}}

\end{table}

\begin{table}[t]\centering
\tiny
\setlength{\tabcolsep}{1.75mm}{

\caption{Generation performance (\%) evaluation on $\mathbb{CARE}$ using Top-3 Documents for Retrieval. 
} 
\label{tab:overall_generation_result}
\begin{tabular}{c|c|c|c|c|c|c|c}
\toprule
 {Retrieval} & {Generation}
& {$\mathbb{R}_c\uparrow $}
& {$\mathbb{R}_i\downarrow$}
& {\textcolor{gray}{$\mathbb{R}_n$}}
& {$\mathbb{N}_c\uparrow $}  
& {$\mathbb{N}_i\downarrow$}    
& {\textcolor{gray}{$\mathbb{N}_n$}}  \\ 
\midrule
\multirow{3}{*}{BM25}
& {Qwen2.5$_{\textsc{7B}}$} 
& {35.7} & {8.23} & {53.4} 
& {3.02} & {0.62} & {4.72} 
\\ 
{} &
{Qwen2.5$_{\textsc{14B}}$} 
& {54.6} & {6.97} & {35.3} 
& {4.39} & {0.55} & {3.01} 
\\ 
{} &
{Qwen2.5$_{\textsc{32B}}$} 
& {57.5} & {6.31} & {33.2} 
& {4.62} & {0.50} & {2.80} 
\\  
\midrule
\multirow{3}{*}{GTE}
& {Qwen2.5$_{\textsc{7B}}$} 
& {34.5} & {7.70} & {55.2} 
& {3.16} & {0.72} & {5.03} 
\\ 
{} &
{Qwen2.5$_{\textsc{14B}}$} 
& {53.3} & {7.02} & {36.3} 
& {4.56} & {0.60} & {3.26} 
\\  
{} &
{Qwen2.5$_{\textsc{32B}}$} 
& {55.3} & {6.31} & {35.3} 
& {4.65} & {0.52} & {3.09} 
\\ 
\midrule
\multirow{3}{*}{BGE-M3}
& {Qwen2.5$_{\textsc{7B}}$} 
& {36.2} & {8.55} & {51.3} 
& {3.25} & {0.72} & {5.06} 
\\  
{} &
{Qwen2.5$_{\textsc{14B}}$} 
& {54.8} & {6.93} & {34.9} 
& {4.48} & {0.56} & {2.97} 
\\ 
{} &
{Qwen2.5$_{\textsc{32B}}$} 
& {57.3} & {6.89} & {32.6} 
& {4.75} & {0.56} & {2.83} 
\\ 
\midrule
\multirow{3}{*}{jina}
& {Qwen2.5$_{\textsc{7B}}$} 
& {38.4} & {\textbf{6.96}} & {51.5} 
& {3.50} & {0.62} & {5.02} \\ 
{} &
{Qwen2.5$_{\textsc{14B}}$} 
& {55.5} & {6.67} & {34.6} 
& {4.53} & {0.54} & {3.01} \\ 
{} &
{Qwen2.5$_{\textsc{32B}}$} 
& {57.0} & {6.87} & {33.0} 
& {4.65} & {0.57} & {2.84} \\ 
\midrule
\multirow{3}{*}{{\model}}
& {Qwen2.5$_{\textsc{7B}}$} 
& {\textbf{60.9}} 
& {7.65} 
& {27.5} 
& {\textbf{4.71}} 
& {\textbf{0.52}} 
& {2.23} \\ 
{} &
{Qwen2.5$_{\textsc{14B}}$} 
& {\textbf{69.2}} 
& {\textbf{4.80}} 
& {22.0} 
& {\textbf{5.32}} 
& {\textbf{0.36}} 
& {1.72} \\ 
{} &
{Qwen2.5$_{\textsc{32B}}$} 
& {\textbf{71.6}} 
& {\textbf{5.40}} 
& {20.5} 
& {\textbf{5.39}} 
& {\textbf{0.38}} 
& {1.53} \\ 
\bottomrule
\end{tabular}}

\end{table}

\subsubsection{Main Results}
From the empirical results on retrieval and generation tasks (Table \ref{tab:overall_retrieval_result_top3} and Table \ref{tab:overall_generation_result}), we can summarize the major findings as follows:
\begin{itemize}[leftmargin=*]
\item \textbf{{\model} largely improves the performance of retrieval and generation for knowledge-intensive tasks.} We only list the retrieval results due to the fact that refusal rate is high when without retrieval (\eg for $\mathbb{INQUIRY}$ the refusal rate is as high as 59.7\% for Qwen2.5$_{\textsc{7B}}$)). By comprehensively combining retrieval and generation metrics defining the correct count, correct ratio, incorrect count, incorrect ratio, we can find that {\model} consistently performs well in different tasks and metrics.
\item \textbf{Both time-evolving and authoritative-sensitive tasks benefit more from {\model}.} A large margin of improvement can be found in $\mathbb{POLICY}$ as it is more sensitive to timeliness and authoritativeness compared to task such as $\mathbb{CARE}$, which depends more on the authoritativeness since the improvement of the treatment takes a lot of time.
\item \textbf{More customization of {\model} \wrt downstream tasks deserves more attention.} We take a trivial step to assign weights to tasks in {\benchmark}, however, the ablation study demonstrates the importance of different views varies across different tasks, hence more attention should be devoted to its customization since each task comes with areas of emphasis.
\end{itemize}

\subsubsection{Feasibility Analysis and Broader Impact}\label{sec:model_discussions}
For industrial platform that directly serves user queries, low-latency inference is of great significance.
In {\model}, we utilize polyviews for a more comprehensive way of information integration that incorporate multiple models in this progress, where the overall procedure is illustrated in the upper part of Figure \ref{fig:framework}. By flexibly incorporating multiple small-scale models and the concurrency and GPU Segmentation mechanisms, the polyview-based integration stage can be deployed using a L20 GPU with latency around 200ms given an user query with an average of 15 documents where the total length exceeds 8k tokens. 
Besides medical applications, for the broader application, the idea of {\model} can also be applied to other domains such as finance where the authoritativeness and timeliness of information greatly matters.
\section{Conclusion and Future Work}\label{sec:con}
In this work, we propose {\model} that incorporates varied views for evaluation of each retrieved document and then pursues integrated polyviews via a multi-reward based view-mixture mechanism, which finally incorporates the derived polyview-grounded knowledge for answer generation. To bridge the evaluation gap we also propose {\benchmark}, a benchmark consists of queries and documents collected from real-world medical scenarios with multiple annotation on them.
Experiments and analysis on {\benchmark} have demonstrated the superiority of {\model}.
Nevertheless, we take a trivial step for the multi-rewards mixture and more complicated approaches requires further research.
In the future, we would like to explore multi-modal retrieval integration and apply the proposed {\model} to other scenarios such as finance.

\clearpage
\bibliography{anthology,custom}
\bibliographystyle{acl_natbib}

\clearpage
\appendix
\section{Appendix}

\subsection{Prompt Template}\label{app:prompt}
This section presents the prompt templates used during training, inference, and evaluation in our proposed {\model}\footnote{Note that since our primary application scenario involves the Chinese language, the initial prompts are provided in Chinese. For your convenience and reference, each prompt template has been translated into English.}.

\subsubsection{Model Training Prompt}
\noindent{\bf Utility Training Prompt.} When training the utility model, we design different prompts so that an LLM can output its perplexity as our supervision signal for embedding model under following circumstances: 
i) answering the question with retrieved document, which demonstrates the utility of the document towards the input question; 
ii) answering the question directly, which means that if the perplexity for answering this question is lower than the perplexity when with retrieved document, then the retrieved document is considered to be useless by the LLM and it could be utilized to achieve selective retrieval. 
Here we present the prompts in Table \ref{tab:utility_w_wo}.
\begin{myverbbox}[\small]{\VerbContentWithInfoD}
Please answer the question based on the given context. Question: 
[QUESTION] The context related to the question is as follows:
[CONTEXT]. Answer: [ANSWER] 
\end{myverbbox}
\begin{myverbbox}[\small]{\VerbContentNoInfoD}
Please answer the question. Question: [QUESTION] Answer: [ANSWER]
\end{myverbbox}

\begin{table}[H]
	\centering
	\caption{Utility Model Training Prompt.}
	\label{tab:utility_w_wo}
	\resizebox{\linewidth}{!}{\begin{tabular}{l}
	\toprule
        w/ retrieved documents \\ \midrule
        \VerbContentWithInfoD       \\ \midrule
        w/o retrieved documents \\ \midrule
        \VerbContentNoInfoD       \\ \midrule
	\end{tabular}}
\end{table}

\noindent{\bf Relevance Training and Inference Prompt.} We evaluate the relevance of the retrieved document \wrt the input query by prompting an LLM with few-shot demonstrations and present the prompts in Table \ref{tab:prompt_relevance}.
\begin{myverbbox}[\small]{\VerbContentWithInfoD}
Your task is to assess the degree of relevance between the Content and the 
Query. The Query consists of a user's question, and the Content contains the 
title and some excerpts from a webpage retrieved online. These Queries and 
Content mainly involve medical knowledge and medical insurance knowledge.

Below are some examples. After reading these examples, I will give you a 
Query and Content. Please assess the relevance of the Content in answering 
the Query and assign a score between A-E (A represents that the Query can be 
fully answered directly by referencing the Content. B represents that the 
Query can still be answered directly by the Content, but the Content contains 
some redundant information or lacks minor details. C represents that the 
Query cannot be directly answered by the Content, but there’s some degree of 
relevance. D represents that the Content cannot directly answer the Query and 
contains only scattered keywords related to the Query. E represents that the 
Content cannot answer the Query at all, and the Content is either meaningless 
or off-topic).

<omited examples>
Example 3:
Query: Pediatric massage
Content: Which department should a child with unexplained fever see? 
Pediatric internal medicine or a fever clinic.
Judge: E
<omited examples>

Now I will provide a Query and Content. Please strictly adhere to the Judge 
format above when providing your judgment and avoid outputting any additional 
content.
Query:{QUESTION}
Content:{CONTEXT}
\end{myverbbox}

\begin{table}[t]
	\centering
	\caption{Relevance Training and Inference Prompt.}
	\label{tab:prompt_relevance}
	\resizebox{\linewidth}{!}{\begin{tabular}{l}
	\toprule
        \VerbContentWithInfoD \\ \midrule
	\end{tabular}}
\end{table}

\noindent{\bf Supplement Training and Inference Prompt.} We evaluate the supplement of the retrieved document \wrt the input query by prompting an LLM with few-shot demonstrations and present the prompts in Table \ref{tab:prompt_supplement}.
\begin{myverbbox}[\small]{\VerbContentWithInfoD}
Your task is to determine whether a piece of Content can serve as 
supplementary information to aid in answering a Query. The Query consists of 
a user's question, and the Content contains the title and some excerpts from 
a webpage retrieved online. These Queries and Content mainly involve medical 
knowledge and medical insurance knowledge.

Regarding supplementary information, here’s a description of the distinction 
between "supplementary information" and "direct answers," using "how to treat 
diabetes" as an example:
(1) Directly answering the Query: Information is considered unable to 
directly answer the Query if the retrieved data is entirely irrelevant or 
provides little to no help in answering "how to treat diabetes." For 
instance, if a user asks about diabetes treatment methods and the returned 
information describes the definition, causes of diabetes, or completely 
unrelated health advice (e.g., general fitness tips that are not specifically 
tailored for diabetic patients), these details cannot help the user 
understand how to treat diabetes and would therefore be deemed irrelevant.
(2) Supplementary information: On the other hand, Content that "provides 
supplementary information" may not directly answer "how to treat diabetes," 
but could contribute additional knowledge, context, or alternative approaches 
that help the user better understand the treatment process or make a more 
informed decision. Examples include:
i. Diet recommendations: Introducing dietary plans for people with diabetes, 
which, while not pharmacological treatments, are critical for managing blood 
sugar levels.
ii. Lifestyle changes: Providing advice on moderate exercise, smoking 
cessation, or limiting alcohol intake, which are beneficial for diabetes 
management.
iii. Psychological support: Discussing mental health maintenance for diabetic 
patients, which, while not a direct physiological treatment, is essential for 
overall patient well-being.
Although such information does not explicitly list specific treatment steps 
or medications, it plays an important role in providing users with a broader 
perspective and support in diabetes management.

In short, whether information is deemed "irrelevant" or "providing 
supplementary information" depends on whether it positively aids the user in 
understanding, deciding, or carrying out actions related to the core question 
(e.g., diabetes treatment). Even indirect information that facilitates the 
user in achieving their query objective can be regarded as supplementary.

Below are some examples. After reading these examples, I will give you a 
Query and Content. Please assess the degree to which the Content provides 
supplementary information for answering the Query and assign a score of 0/1 
(1 represents that the Content provides supplementary information, while 0 
represents that it does not provide supplementary information).

Example 1:
Query: How to reverse mild fatty liver disease?
Content: What are the stages of fatty liver disease? Simple steatosis: 
Symptoms include fatigue and upper right abdominal discomfort, with normal 
liver function. Ultrasound or (and) CT scans indicate mild to moderate fatty 
liver. Steatohepatitis: Symptoms include fatigue and upper right abdominal 
discomfort, with liver function exceeding the upper normal limit by 1-5 times 
for over four weeks. Ultrasound or (and) CT scans indicate fatty liver. 
Hepatic fibrosis or (and) cirrhosis: Symptoms include fatigue and upper right 
abdominal discomfort, with liver function and blood indicators of fibrosis 
being normal or abnormal. Ultrasound or (and) CT, MRI, liver stiffness 
testing, etc., suggest fatty liver with fibrosis or cirrhosis confirmed by 
liver biopsy.
Judge: 1
<omitted examples>

Now I will provide a Query and Content. Please strictly adhere to the Judge 
format above when providing your judgment and avoid outputting any additional 
content.
Query:{QUESTION}
Content:{CONTEXT} 
\end{myverbbox}

\begin{table}[t]
	\centering
	\caption{Supplement Training and Inference Prompt.}
	\label{tab:prompt_supplement}
	\resizebox{\linewidth}{!}{\begin{tabular}{l}
	\toprule
        \VerbContentWithInfoD \\ \midrule
	\end{tabular}}
\end{table}

\subsubsection{Generation Stage Prompt}
To prompt an LLM such that it can generate output for domains $\mathbb{INQUIRY}$, $\mathbb{POLICY}$  and $\mathbb{CARE}$ as we required, we utilize the different prompts when (not) incorporating retrieved documents in different domains and the detailed prompts can be found in Table \ref{tab:prompt_inquiry}, Table \ref{tab:prompt_policy} and Table \ref{tab:prompt_care}.

\begin{myverbbox}[\small]{\VerbContentWithInfoD}
system:
Please answer the following question based on the "Reference Materials," 
adhering to the requirements below:
1. Provide an answer that is as concise, polite, and logical as possible, 
under 300 words.
2. Use the "general-specific-general" format and markdown structure in 
your response.
3. If it is not possible to answer based on the content in the Reference 
Materials, reply with: "Sorry, I do not have the relevant knowledge yet."
4. Do not forget that you are a medical assistant. Offer positive and 
constructive advice or educational explanations related to the issue 
without providing definitive diagnostic opinions like a doctor.
5. Do not use <|Reason|> to start your reasoning. Begin your final answer 
with the tag <|ANSWER|> and end your response in the format <|ANSWER|>: 
\$answer.

user:
Question:
[QUESTION]
Reference Materials
[CONTEXTS]
\end{myverbbox}
\begin{myverbbox}[\small]{\VerbContentNoInfoD}
system:
Please answer the following questions with the following requirements:
1. Provide answers that are as concise, polite, logical, and under 300 
words as possible.
2. Use the "general-specific-general" structure and markdown format for 
answering.
3. If unable to answer, respond with: "Sorry, I do not have the relevant 
knowledge yet."
4. Do not forget that you are a medical assistant. Offer positive and 
constructive advice or scientific explanations related to the issue 
without providing definitive diagnostic opinions like a doctor.
5. Do not begin thinking with <|Reason|>; instead, start your final 
answer with the tag <|ANSWER|> and conclude your reply in the format 
<|ANSWER|>: \$answer.

user:
Question:
[QUESTION]
\end{myverbbox}

\begin{table}[t]
	\centering
	\caption{Generation Prompt for $\mathbb{INQUIRY}$.}
	\label{tab:prompt_inquiry}
	\resizebox{\linewidth}{!}{\begin{tabular}{l}
	\toprule
        w/ retrieved documents \\ \midrule
        \VerbContentWithInfoD       \\ \midrule
        w/o retrieved documents \\ \midrule
        \VerbContentNoInfoD       \\ \midrule
	\end{tabular}}
\end{table}

\begin{myverbbox}[\small]{\VerbContentWithInfoD}
system: 
Please answer the question based on the "Reference Materials" with the 
following requirements:
1. Ensure that your response is polite, logical, and no more than 300 
words.
2. If the answer requires providing detailed steps, include all details 
as mentioned in the original text, and do not omit any steps.
3. If the reference materials mention specific regions, do not omit them 
in your response. You can specify by saying “For example, in [region].”
4. Avoid using terms like "New Rural Cooperative Medical Scheme" (also 
called NCMS, cooperative medical care, rural cooperative healthcare, or 
rural medical insurance), as they no longer exist. Inform users that it 
has been merged into the Urban and Rural Resident Basic Medical Insurance.
5. Do not begin with <|Reason|> when reasoning. Start your final answer 
with the tag <|ANSWER|> and end your response in the format <|ANSWER|>: \$answer.

user:
Question:
[QUESTION]
Reference Materials
[CONTEXTS]
\end{myverbbox}
\begin{myverbbox}[\small]{\VerbContentNoInfoD}
system:
Please answer the following questions with the requirements below:
1. Ensure that your response is polite, logical, and no more than 300 
words.
2. If you have relevant professional knowledge and there are detailed 
steps available, provide the steps in full without omitting them.
3. If the response requires mentioning specific regions, do not omit the 
locations. You can specify by saying “For example, in [region].”
4. Avoid using terms like "New Rural Cooperative Medical Scheme" (also 
known as NCMS, cooperative medical care, rural cooperative healthcare, or 
rural medical insurance), as they no longer exist. Instead, inform users 
that it has been merged into the Urban and Rural Resident Basic Medical 
Insurance.
5. Do not begin with <|Reason|> when reasoning. Start your final answer 
with the tag <|ANSWER|> and end your response in the format <|ANSWER|>: \$answer.

user:
Question:
[QUESTION]
\end{myverbbox}

\begin{table}[t]
	\centering
	\caption{Generation Prompt for $\mathbb{POLICY}$.}
	\label{tab:prompt_policy}
	\resizebox{\linewidth}{!}{\begin{tabular}{l}
	\toprule
        w/ retrieved documents \\ \midrule
        \VerbContentWithInfoD       \\ \midrule
        w/o retrieved documents \\ \midrule
        \VerbContentNoInfoD       \\ \midrule
	\end{tabular}}
\end{table}

\begin{myverbbox}[\small]{\VerbContentWithInfoD}
system:
You are a medical expert with professional healthcare knowledge and excel 
at using plain and understandable language to provide educational 
explanations for patients. Please base your answers on the following 
execution steps and respond to the patient's question step by step:

Execution Steps:
1. Understand the patient's question and consider the key information 
points the patient is most eager to learn when asking the question.
2. Think about the specific content that should be included in those key 
information points. You may use your professional knowledge or consult 
the reference materials to answer. If the content from the reference 
materials is incorrect, do not use it. If you lack the relevant 
expertise, reply with: "Sorry, I do not have the relevant knowledge yet."
3. Organize the information from steps 1 and 2 logically, such as by 
using categorization or progressive relationships.
4. Provide a comprehensive and logical answer, and include a risk warning 
at the end to help avoid potential medical disputes.
5. For "yes or no" type questions, clearly state your conclusion upfront, 
such as: "Yes," "Not recommended," or "No."
6. If the patient’s condition appears to be dangerous, advise the patient 
to seek medical attention promptly.

Output Requirements:
1. Use plain and simple language, avoiding overly technical terms.
2. Keep the response brief but thorough, with a clear and easy-to-read 
format. Do not omit key points, avoid wordiness, and ensure brevity, as 
users may not have the patience for lengthy responses.
3. Answers must adhere to medical facts; no fabricated information is 
allowed.
4. Provide only the final answer; do not display your reasoning process.
5. The response should not exceed 250 words.

user:
Question:
[QUESTION]
Reference Materials
[CONTEXTS]
\end{myverbbox}
\begin{myverbbox}[\small]{\VerbContentNoInfoD}
system:
You are a medical expert with professional healthcare knowledge and excel 
at using plain and understandable language to provide educational 
information to patients. Please base your answers on the following 
execution steps and answer the patient's question step by step:

Execution Steps:
1. Understand the patient's question and consider the key information 
points the patient is most eager to learn when asking the question.
2. Think about the specific content that should be included in those key 
information points. Use your professional knowledge to answer; if you 
lack the relevant knowledge, respond with "Sorry, I do not have the 
relevant expertise.”
3. Organize the information from steps 1 and 2 logically, such as using 
categorization or progressive relationships.
4. Provide a comprehensive and logical answer, and include a risk warning 
at the end of the answer to help avoid medical disputes.
5. For "yes or no" type questions, clearly state your conclusion upfront, 
e.g., "Yes," "Not recommended," or "No."
6. For situations where the patient's condition may be dangerous, suggest 
that they seek medical attention promptly.

Output Requirements:
1. Use plain and simple language, avoiding overly technical terms.
2. Keep the response brief but thorough, with a clear and easy-to-read 
format. Avoid omitting key points or being excessively wordy, as users 
may not have the patience to read overly long responses.
3. Answers must align with medical facts; absolutely no fabricated 
information is allowed.
4. Provide the final answer only; do not display your thinking process.
5. The overall response should not exceed 250 words.

user:
Question:
[QUESTION]
\end{myverbbox}

\begin{table}[t]
	\centering
	\caption{Generation Prompt for $\mathbb{CARE}$.}
	\label{tab:prompt_care}
	\resizebox{\linewidth}{!}{\begin{tabular}{l}
	\toprule
        w/ retrieved documents \\ \midrule
        \VerbContentWithInfoD       \\ \midrule
        w/o retrieved documents \\ \midrule
        \VerbContentNoInfoD       \\ \midrule
	\end{tabular}}
\end{table}

\subsubsection{Auto-evaluation Prompt}
We evaluate each generation result by incorporating GPT4 as the judge model, we first generate different statements in the answer (please refer to Table \ref{tab:prompt_asg} for details) and then check the ratio of statements of the generation result that has been correctly mentioned in the ground truth from human experts (please refer to Table \ref{tab:prompt_asj} for details).

\begin{myverbbox}[\small]{\VerbContentWithInfoD}
[Instruction] You are a medical insurance expert. Given a question and an 
answer, generate one or more factual statements from each sentence of the 
answer. 

[Requirements]:
The generated statements must not contain pronouns. If necessary, pronouns 
can be rewritten using the overall context of the answer or the question.
The generated statements must be complete. If necessary, the cause and effect 
can be supplemented based on the context.
The generated statements must be entirely derived from the answer and must 
not alter the original meaning.
If a specific procedure is mentioned, the entire procedure must be included 
in one statement. If there are multiple procedures, they must be included in 
separate statements.

[Here is an example]:
Question
How can I use my medical insurance balance for my family members who are part 
of the shared account?

Answer
To use your medical insurance balance for your family members who are part of 
the shared account, you can follow these steps:

Setting up Family Sharing:
First, you need to set up the family sharing binding.
On the Alipay homepage, click on [Medical Health] or search for “Medical 
Health,” enter the Alipay [Medical Health] mini-program, search for [Family 
Sharing], click [Use Now], click [Apply Now], and follow the operation 
prompts to complete the setup.

Using the Medical Insurance Electronic Voucher:
During payment, display the medical insurance electronic voucher for scanning 
and settlement.
The system will prioritize deducting from the balance in your account.
When your account balance is insufficient, the system will automatically use 
the personal account balance of the family members in the shared account.

Handling Special Cases:
For infants or elderly family members without mobile devices, the family 
member can use the family account feature in the National Medical Insurance 
Bureau APP to display the electronic voucher for settlement.
Please note that the use and management of family sharing funds must comply 
with local medical insurance regulations. Violating these regulations will 
result in corresponding legal liabilities.

Statements
To use the medical insurance balance for family members, first, set up the 
family sharing binding.
The procedure for setting up family sharing is: On the Alipay homepage, click 
on [Medical Health] or search for “Medical Health,” enter the Alipay [Medical 
Health] mini-program, search for [Family Sharing], click [Use Now], click 
[Apply Now], and follow the operation prompts to complete the setup.
When using the medical insurance balance for family members, display the 
medical insurance electronic voucher for scanning and settlement.
When using the medical insurance balance for family members, the system 
prioritizes deducting from the balance in the account.
When using the medical insurance balance for family members, if the account 
balance is insufficient, the system will automatically use the personal 
account balance of the family members in the shared account.
When using the medical insurance balance for family members, if there are 
special cases such as infants or elderly family members without mobile 
devices, the family member can use the family account feature in the National 
Medical Insurance Bureau APP.
[Please generate the following results based on the requirements and example]:

Question
${QUESTION}

Answer
${ANSWER}

Statements
\end{myverbbox}

\begin{table}[t]
	\centering
	\caption{Answer Statement Generation Prompt.}
	\label{tab:prompt_asg}
	\resizebox{\linewidth}{!}{\begin{tabular}{l}
	\toprule
        \VerbContentWithInfoD \\ \midrule
	\end{tabular}}
\end{table}

\begin{myverbbox}[\small]{\VerbContentWithInfoD}
[Instruction] You are an expert in the field of medical insurance. 
Considering the given question, the real answer, and multiple statements, 
judge whether each statement is incorrect, not mentioned, or correct, and 
provide the reason.

[Requirements]:
1. Combine the question to understand the overall meaning of the real 
answer, understand each reference relationship in the answer, and understand 
each logical relationship of and, or, not, before judging each statement.
2. The criteria for judging "not mentioned" are as follows: 
2.1 If the argument mentioned in the statement does not exist in the real 
answer or cannot be inferred, it is considered not mentioned. 
2.2 If the statement answers from multiple perspectives, but the real answer 
only covers one perspective, it is considered not mentioned. 
2.3 If the correctness of the statement cannot be verified based on the real 
answer, it is considered not mentioned.
3. The criteria for judging "incorrect" are as follows: 
3.1 If the statement mentions "related app," "related application," "medical 
insurance app," or other vague expressions, it is considered incorrect. 
3.2 If the argument mentioned in the statement is also mentioned or can be 
inferred from the real answer, and you can verify that the argument in the 
statement is incorrect using the real answer, it is considered incorrect. If 
you cannot prove the argument is incorrect based on the real answer, do not 
consider it incorrect. 
3.3 For statements about the process, only judge that the process exists in 
the real answer. It is considered incorrect only when the process does not 
exist in the real answer.
4. The criteria for judging "correct" are as follows: 
4.1 If the argument in the statement is also mentioned or can be inferred 
from the real answer, and there is no contradiction, it is considered 
correct. 
4.2 If none of the situations in 2 and 3 apply, it is considered correct.
After indicating the judgment result with "not mentioned" / "incorrect" / 
"correct," use a semicolon to separate the reason.

[Here is an example]:
Question
How can I use my medical insurance balance for my family members who are part 
of the shared account?

Answer
To use your medical insurance balance for your family members who are part of 
the shared account, you can follow these steps:
1. Set up Family Sharing:
On the Alipay homepage, click on [Healthcare] or search for “Healthcare,” 
enter the Alipay [Healthcare] mini-program, search for [Family Sharing], 
click [Use Now], click [Apply Now], and follow the prompts to complete the 
setup.
2. Use the Electronic Medical Insurance Card:
When making a payment, show the electronic medical insurance card for 
scanning.
The system will prioritize deducting from the balance of the current user's 
electronic medical insurance card.
If the user's account balance is insufficient, it will automatically use the 
personal account balance of the authorized person.
3. Special Case Handling:
For infants or elderly family members without mobile devices, you can use the 
Alipay family account feature to display the user's electronic card to 
complete the transaction.
Please note that the use and management of family sharing funds must comply 
with local medical insurance regulations. Misuse of funds will result in 
corresponding legal responsibilities.

Statements
1. To use for family members, you need to set up family sharing.
3. The setup path is: On the Alipay homepage, click on [Healthcare] or search 
for “Healthcare,” enter the Alipay [Healthcare] mini-program, search for 
[Family Sharing], click [Use Now], click [Apply Now], and follow the prompts 
to complete the setup.
3. When using for family members, you need to show the electronic medical 
insurance card for scanning.
4. When using for family members, the system will prioritize deducting from 
your balance.
5. When using for family members, if your account balance is insufficient, it 
will automatically use the personal account balance of the family member.
6. When using for family members, if there are special cases such as infants 
or elderly family members without mobile devices, you can use the family 
account feature of the National Medical Insurance Bureau app.

Judgment
1. Correct; The real answer mentions following the steps, the first step is 
to set up family sharing, which can be inferred from the statement, and there 
is no contradiction.
2. Correct; The real answer mentions the setup path for family sharing, which 
is consistent with the statement.
3. Correct; The real answer mentions that when using, you need to show the 
electronic medical insurance card for scanning, which is consistent with the 
statement.
4. Incorrect; The real answer mentions that when using, the system 
prioritizes deducting from the user's account balance. Based on the question, 
the user refers to the family member, which is inconsistent with the 
deduction subject mentioned in the statement.
5. Incorrect; The real answer mentions that when using, if the user's account 
balance is insufficient, it will automatically use the personal account 
balance of the authorized person. The user refers to the family member, and 
the authorized person is you, which is opposite to the subject mentioned in 
the statement.
6. Not mentioned; The statement mentions that it can be used through the 
National Medical Insurance Bureau app, but the real answer does not mention 
this, only stating that it can be used through the Alipay app, and it is 
unclear whether the National Medical Insurance Bureau app can be used, so it 
cannot be verified as correct or incorrect, hence it is not mentioned.

Question
${QUESTION}

Real Answer
${GROUNDTRUTH}

Statements
${STATEMENT}

Judgment
\end{myverbbox}

\begin{table}[t]
	\centering
	\caption{Answer Statement Judgement Prompt.}
	\label{tab:prompt_asj}
	\resizebox{\linewidth}{!}{\begin{tabular}{l}
	\toprule
        \VerbContentWithInfoD \\ \midrule
	\end{tabular}}
\end{table}



\clearpage
\end{document}